\documentclass[conference]{IEEEtran}
\usepackage{cite}
\usepackage{amsmath,amsfonts}
\usepackage{graphicx}
\usepackage{textcomp}
\usepackage{color}
\usepackage{colortbl}
\usepackage{multirow}
\usepackage[T1]{fontenc}
\usepackage[utf8]{inputenc}
\usepackage{xspace}
\usepackage{latexsym} 
\usepackage[usenames,dvipsnames]{pstricks}
\usepackage{epsfig}
\usepackage{pst-plot}

\usepackage{array}            
\usepackage{multicol}         
\usepackage{graphicx}         
\usepackage{latexsym}         
\usepackage{amsthm}
\usepackage{systeme}

\usepackage{url}

\usepackage{subfig}
\usepackage{epstopdf}
\usepackage{arydshln}
\usepackage{pgf,tikz,pgfplots}
    
\usepackage{algorithm}
\usepackage[noend]{algpseudocode}
\usetikzlibrary {positioning}
\usepackage{fancyvrb}

\usepackage{balance}

\definecolor {processgray}{cmyk}{0,0,0,0.5}

\newcommand{\instance}[0]{\ensuremath{\gamma}\xspace}
\newcommand{\batch}[0]{\ensuremath{{\bf B}}\xspace}
\newcommand{\normalization}[0]{\ensuremath{\mathbf{Z}}\xspace}

\newcommand{\numberOfRealisations}[0]{\ensuremath{{n_r}}\xspace}
\newcommand{\incBetaFunc}[3]{\ensuremath{\mathbf{I}}_{#3}(#1,#2)\xspace}
\newcommand{\sample}[4]{\ensuremath{\text{Sample}}_{#4}(#1, #2, #3)\xspace}
\newcommand{\reservoirSize}[0]{\ensuremath{k}\xspace}
\newcommand{\datasize}[0]{\ensuremath{n}\xspace}
\newcommand{\reservoir}[0]{\ensuremath{\mathcal{S}}\xspace}

\newcommand{\specialization}[0]{\ensuremath{\preceq}\xspace}
\newcommand{\uFunction}[1]{\ensuremath{F_{#1}}\xspace}
\newcommand{\lengthOrNorm}[0]{\ensuremath{\ell}\xspace}
\newcommand{\temporalbiasfunction}[2]{\ensuremath{\nabla_{\!#1}(#2)}\xspace}
\newcommand{\insertPattern}[0]{\ensuremath{\triangleright}\xspace}

\newcommand{\timestamp}[0]{\ensuremath{t}\xspace}

\newcommand{\expo}[1]{\ensuremath{e^{#1}}\xspace}
\newcommand{\dampingFactor}[0]{\ensuremath{\varepsilon}\xspace}

\newcommand{\acceptanceProb}[0]{\ensuremath{p}\xspace}

\newcommand{\sequenceOf}[1]{\ensuremath{\langle #1 \rangle}\xspace}

\newcommand{\alphaDecay}[0]{\ensuremath{\alpha}\xspace}
\newcommand{\minNorm}[0]{\ensuremath{\mu}\xspace}
\newcommand{\maxNorm}[0]{\ensuremath{M}\xspace}

\newcommand{\mFreq}[0]{\ensuremath{f\!req}\xspace}
\newcommand{\mArea}[0]{\ensuremath{area}\xspace}
\newcommand{\mInvArea}[0]{\ensuremath{area^{-1}}\xspace}
\newcommand{\mDecay}[0]{\ensuremath{decay}\xspace}
\newcommand{\mUtility}[0]{\ensuremath{hui}\xspace}
\newcommand{\mAvgUtility}[0]{\ensuremath{haui}\xspace}
\newcommand{\mMinNorm}[0]{\ensuremath{\geq \minNorm}\xspace}
\newcommand{\mMaxNorm}[0]{\ensuremath{\leq\maxNorm}\xspace}

\newcommand{\items}[0]{\ensuremath{\mathcal{I}}\xspace}

\newcommand{\lang}[0]{\ensuremath{\mathcal{L}}\xspace}
\newcommand{\database}[0]{\ensuremath{\mathcal{D}}\xspace}
\newcommand{\literal}[0]{\ensuremath{e}\xspace}
\newcommand{\card}[1]{\ensuremath{{\left|{#1}\right|}}\xspace}
\newcommand{\weight}[0]{\ensuremath{\omega}\xspace}
\newcommand{\prob}[0]{\ensuremath{\mathbb{P}}\xspace}

\newcommand{\maxLen}[0]{\ensuremath{M}\xspace} 
\newcommand{\minLen}[0]{\ensuremath{\mu}\xspace} 

\newcommand{\uLen}[0]{\ensuremath{\ell}\xspace}
\newcommand{\measure}[0]{\ensuremath{m}\xspace}

\newcommand{\pattern}[0]{\ensuremath{\varphi}\xspace} 
\newcommand{\complexity}[0]{\ensuremath{O}\xspace}  

\newcommand{\globalUtilMeasure}[0]{\ensuremath{G}\xspace}

\newcommand{\itemset}[0]{\ensuremath{X}\xspace}

\newcommand{\langS}[0]{\ensuremath{\lang_{\text{S}}}\xspace}
\newcommand{\langI}[0]{\ensuremath{\lang_{\text{I}}}\xspace}

\newcommand{\norm}[1]{\ensuremath{\|#1\|}\xspace}

\newcommand{\utilityTrans}[0]{\ensuremath{u}\xspace}
\newcommand{\utility}[0]{\ensuremath{U}\xspace}

\newcommand{\myalgo}[0]{\ensuremath{{\textrm{\sc {\small RPS}}}}\xspace}

\newcommand{\MLPClassifier}[0]{\ensuremath{{\textrm{\sc MLP}}}\xspace}
\newcommand{\SGDClassifier}[0]{\ensuremath{{\textrm{\sc SGD}}}\xspace}
\newcommand{\PassiveAggressiveClassifier}[0]{\ensuremath{{\textrm{\sc PAC}}}\xspace}
\newcommand{\Perceptron}[0]{\ensuremath{{\textrm{\sc Per}}}\xspace}
\newcommand{\MultinomialNB}[0]{\ensuremath{{\textrm{\sc MNB}}}\xspace}

\newcommand{\respat}[0]{\ensuremath{{\textrm{\sc ResPat}}}\xspace}

\newtheorem{definition}{Definition}
\newtheorem{example}{Example}
\newtheorem{property}{Property}

\newtheorem*{problem}{Problem Definition}

\begin{document}
\title{
RPS: A Generic Reservoir Patterns Sampler
}%

 \author{
     \IEEEauthorblockN{Lamine Diop\IEEEauthorrefmark{1}, Marc Plantevit\IEEEauthorrefmark{1}, Arnaud Soulet\IEEEauthorrefmark{2}}
     \IEEEauthorblockA{
         \IEEEauthorrefmark{1}EPITA Research Laboratory (LRE), Le Kremlin-Bicetre, Paris FR-94276, France, 
         firstname.lastname@epita.fr
     }
    \IEEEauthorblockA{
        \IEEEauthorrefmark{2}University of Tours, LIFAT, 3 Place Jean Jaurès, Blois 41029, France, arnaud.soulet@univ-tours.fr
    }
}

\maketitle

\begin{abstract}
Efficient learning from streaming data is important for modern data analysis due to the continuous and rapid evolution of data streams. Despite significant advancements in stream pattern mining, challenges persist, particularly in managing complex data streams like sequential and weighted itemsets. While reservoir sampling serves as a fundamental method for randomly selecting fixed-size samples from data streams, its application to such complex patterns remains largely unexplored. 
In this study, we introduce an approach that harnesses a weighted reservoir to facilitate direct pattern sampling from streaming batch data, thus ensuring scalability and efficiency. We present a generic algorithm capable of addressing temporal biases and handling various pattern types, including sequential, weighted, and unweighted itemsets. Through comprehensive experiments conducted on real-world datasets, we evaluate the effectiveness of our method, showcasing its ability to construct accurate incremental online classifiers for sequential data. Our approach not only enables previously unusable online machine learning models for sequential data to achieve accuracy comparable to offline baselines but also represents significant progress in the development of incremental online sequential itemset classifiers.
\end{abstract}

\begin{IEEEkeywords}
Reservoir sampling, Output pattern sampling
\end{IEEEkeywords}

\section{Introduction}
\label{sec:Introduction}

Stream data mining is a subset of data mining, aiming to extract valuable knowledge, patterns, and insights from continuously flowing data streams \cite{AGGARWAL200381}. Unlike static data, data streams consist of an unbounded, constant flow of information from diverse sources such as sensors, social media, financial transactions, and network traffic. Various applications and algorithmic advancements have emerged in stream data mining. Sequential pattern mining, for example, is crucial for market basket analysis and web clickstream analysis \cite{zhang2023effective}. Additionally, efficient algorithms have facilitated real-time analytics in anomaly detection, retail analysis \cite{toliopoulos2020continuous}, probabilistic neural networks \cite{rutkowska2023l2convergence}, and high utility itemsets in weighted itemsets \cite{li2023fchmstream}. Moreover, methods for mining periodic batches and detecting drift \cite{liu2023hypercalm} have also been proposed.

Despite these successes, stream data mining, characterized by its continuous and rapidly changing nature, poses unique challenges for traditional data processing techniques. Reservoir sampling has emerged as a fundamental method for randomly selecting a fixed-size sample from data streams, offering simplicity and constant space complexity. Reservoir pattern sampling has been recently proposed \cite{giacometti2021reservoir} by adapting the reservoir sampling approach \cite{efraimidis2006weighted} for itemset only. However, despite these advancements, challenges persist. These techniques may face limitations when handling large and rapidly evolving complex structured data such as sequential itemsets \cite{srikant1996mining} or weighted itemsets \cite{Tseng2016}. These limitations underscore the ongoing need for innovative approaches to address the evolving complexities of stream data mining.

To overcome these challenges, we introduce an extension of the multi-step pattern sampling technique \cite{BLPG11,diopICDM18,DiopPAKDD2022} tailored for stream data. Despite its success, applying multi-step pattern sampling in complex and structured data streams remains unexplored. By leveraging a weighted reservoir, our approach enables the direct sampling and maintenance of patterns from streaming batch data, offering scalability and efficiency. We present a generic algorithm capable of handling temporal biases and various pattern types, such as sequential, weighted and unweighted itemsets, and discuss its effectiveness in addressing the long-tail issue commonly encountered in pattern sampling tasks. We also show the usefulness of the sampled patterns by proposing online classifiers on stream sequential itemsets with many models that were not able to run with sequential data. 

The primary contributions of this paper include:

\begin{itemize}
    \item We propose the first reservoir pattern sampling approach for complex structured data such as sequential and weighted itemsets in stream batches. Using the multi-step technique, we present a fast pattern sampling approach that leverages the inverse incomplete Beta function and efficient computation of the normalization constant.
    \item Our algorithm named \myalgo is generic and works with temporal biases such as damped window and landmark windows while integrating numerous interestingness measures like frequency, area, and decay combined with any norm-based utility to avoid the long-tail problem where long and rare patterns flood the space. We also present a large set of experimental results for analyzing the behavior of \myalgo with diverse types of parameters.
    \item We show the usefulness of the sampled patterns for online classifier building for sequential data classification with new labels arrival. Specifically, we adapt several classification models  for online sequential data classification with unseen labels, which, to the best of our knowledge, is a novel contribution. Experimental results indicate that sampled patterns are highly effective in constructing accurate classifiers for online sequential data.
    \item  For reproducibility and open science purpose, the source code and the experiments are made available on a public repository\footnote{The code is available at: \url{https://github.com/RPSampler/RPS}.}.
\end{itemize}

The structure of this paper is organized as follows: In Section \ref{sec:Related}, we provide a review of related work concerning reservoir sampling techniques and multi-step pattern sampling methods. Section \ref{sec:Problem} presents the fundamental definitions and formal problem statement. Our proposed generic solution is detailed in Section \ref{sec:contrib}. In Section \ref{sec:Experiments}, we conduct an evaluation of our approach using real-world and benchmark datasets, comparing the accuracy of a sample-based online sequential data classifier with state-of-the-art methods. Finally, we conclude the paper and discuss future directions in Section \ref{sec:Conclusion}.
\section{Related Work}
\label{sec:Related}

This section presents the reservoir sampling in stream data and the local multi-step pattern sampling literature.

\subsection{Reservoir sampling}
Reservoir sampling is a fundamental technique in computer science and statistics used to address the problem of randomly selecting a fixed-size sample from a stream of data without knowing the total number of elements in advance \cite{vitter1985random}. The primary motivation behind reservoir sampling is to efficiently sample elements from large or infinite data streams \cite{alkateb2014adaptive} without lost of soundness, where traditional methods like sorting or storing all the data are impractical due to memory constraints \cite{cormode2010optimal}. One of the key benefits of reservoir sampling is its simplicity and constant space complexity \cite{efraimidis2006weighted}, making it suitable for real-time data processing and applications with limited resources. It is widely used for tasks like estimating statistical properties of large datasets, and sampling representative subsets of data for training machine learning models. Raissi and Poncelet \cite{ChedyICDM07} utilize reservoir sampling for input sampling (\textit{subset of instances from the database}) before mining sequential patterns with bounded error rates for both static databases and data streams.
Recently, it has been extended to output pattern sampling (\textit{subset of patterns from the pattern language}) in stream itemsets \cite{giacometti2021reservoir} where each transaction or itemset \instance is spread into a set of patterns, $2^\instance\setminus\emptyset$, without materialize it. After that, the set of patterns is scanned using an binary index operator to draw a pattern directly. 

However, with a large number of patterns per transaction, the computational complexity of maintaining the reservoir can become a bottleneck. The need to process and sample from an extensive list of patterns within each transaction can slow down the sampling process, making it less efficient. In addition, the reservoir sampling technique proposed in \cite{giacometti2021reservoir} is not scalable because the key idea which based on the binary index operator is not applicable with complex structure such as sequence \cite{zhang2023effective} and quantitative data (weighted itemsets) \cite{li2023fchmstream}.

\subsection{Local multi-step output pattern sampling}
Multi-step pattern sampling \cite{BLPG11,giacometti2018dense} is the fastest among the techniques used in output space pattern sampling \cite{al2009output,BLPG11,dzyuba2017flexible} to draw representative patterns directly from the database. Particularly efficient for sampling in local data, multi-step is widely regarded as the most efficient approach, especially following the preprocessing phase, which involves computing the normalization constant. This method has been successfully applied across different pattern languages, including itemset \cite{BLPG11}, numerical data \cite{giacometti2018dense}, sequential data \cite{diopICDM18}, and quantitative data \cite{DiopPAKDD2022}. The primary concept behind this technique is to draw a pattern directly from the database with a probability proportional to a given interestingness measure \measure. This involves two steps after the preprocessing phase, wherein each instance \instance of the database is weighted by the sum of the total utility of the set of patterns it contains. In the first step, an instance \instance is randomly drawn with a probability proportional to its weight, while the second step allows for the drawing of a pattern proportionally to its utility $\measure(\pattern, \instance)$ from the set of patterns of \instance. However, one of its most intricate limitations is the requirement to know the total sum of utility of the patterns, which can be time-consuming with very large databases or unfeasible with stream data \cite{Kimura2022}.

In this paper, we demonstrate how to extend the multi-step pattern sampling technique to sample and maintain a set of patterns directly from stream data based on a weighted reservoir. We propose a generic algorithm capable of handling itemsets, sequential patterns, and high utility itemsets while incorporating norm-based utility to address the long-tail issue.
\section{Problem Statement}
\label{sec:Problem}

This section formalizes the problem of reservoir-based multi-step pattern sampling under norm-based utility measure. We first recall some preliminary definitions about structured patterns and stream data.

\subsection{Definitions and preliminaries}

Let $\items = \{\literal_1, \ldots, \literal_N\}$ be a set of finite literals called items. An itemset $\itemset$ is a non-empty subset of $\items$, i.e., $\itemset \in 2^\items\setminus\emptyset$. The set of all itemsets in $\items$ is called the pattern language for itemset mining, denoted by $\langI$. 
An instance $\instance = \sequenceOf{\itemset_1, \ldots, \itemset_n}$ defined over $\items$ is an ordered list of itemsets $\itemset_i \in \lang_\items$ $(1 \leq i \leq n, n \in \mathbb{N})$. $n$ is the size of the instance $\instance$ denoted by $\card{\instance}$. If $\card{\instance}>1$ then \instance is a sequence of itemsets and $\langS$ denote the universal set of all sequences defined over $\items$, otherwise \instance is an itemset also called a transaction denote by $\instance=\itemset_1$ for simplicity. A transaction can be weighted and the patterns mined from it are called high utility itemset (HUI) in general. 
High utility itemset is dedicated to itemset discovery from a quantitative database where each item of an instance is associated with a weight, which is a strictly positive real number depending on the instance and referred to as its utility. The norm of an instance $\instance$, denoted by $\norm{\instance}$, is the sum of the cardinality of all its itemsets, i.e., $\norm{\instance} = \sum_{i=1}^n \card{\itemset_i}$. Finally, given a pattern language $\lang \in \{\langI, \langS\}$, pattern $\pattern \in \lang$ can be generally defined as follows:

\begin{definition}[Pattern]
	$\pattern = \sequenceOf{\itemset'_1, \ldots, \itemset'_{n'}}$ is a pattern or an generalization of an instance $\instance = \sequenceOf{\itemset_1, \ldots, \itemset_n}$, denoted by $\pattern \specialization \instance$, if there exists an index sequence $1 \leq i_1 < i_2 < \ldots < i_{n'} \leq n$ such that for all $j \in [1..{n'}]$, one has $\itemset'_j \subseteq \itemset_{i_j}$. 
\end{definition}

This definition is usually used in the context of sequential pattern mining, but we recall that an itemset is nothing else that a sequential pattern of length $1$.

\noindent {{\bf Data stream and interestingness utility measures:}} 
In general, we denote $\lang \in \{\langI, \langS\}$ as a pattern language. A data stream is a sequence of batches with timestamps denoted as follows: $\database = \sequenceOf{(\timestamp_1, \batch_1), \ldots, (\timestamp_n, \batch_n)}$, such that $\batch_j \subseteq \lang$ for all $j \in [1..n]$ and $\timestamp_j < \timestamp_{j+1}$ for all $j \in [1..n-1]$, where a batch is a set of finite instances send at the same time, i.e., $$\batch_j=\{\instance_{j_1}, \cdots, \instance_{j_{j'}}: (\instance_{j_k} \in \lang) (\forall~ k \in [1..j'])\}.$$ In other words, a batch contains a set of instances that have equal temporal relevance. $\lang(\database)$ is the set of all patterns that can be mined from \database. In this paper, we consider the \emph{Landmark window} time constraint, which provides a structured way to divide the data stream into manageable chunks called instances, and \emph{damped window} which favors the recent instances. We also use other constraints and utility measures that can combine frequency and norm-based utility measures.

\begin{definition}[Frequency]
	Given a database $\database$ defined over a pattern language $\lang$, the frequency of a pattern $\pattern \in \lang$ denoted $\mFreq(\pattern, \database)$, is the number of instances that support. Formally, it is defined as follows: $$\mFreq(\pattern, \database) = \card{\{\instance \in \batch: ((\timestamp, \batch) \in \database)\wedge(\pattern \specialization \instance)\}}.$$
\end{definition}

In pattern mining, frequency is often associated with other interestingness measures to reveal meaningful insights. In this paper, we combine it with other measures, specifically norm-based utility measures \cite{diop2019kais}, to identify truly interesting and actionable patterns. It is also possible and helpful to use norm-based utility measures in high utility itemset discovery.

\begin{definition}[Norm-based utility \cite{diop2019kais}]
	\label{def:HUI}
	A utility function $\uFunction{\measure}$ is a norm-based utility if there exists a function $f_\measure : \mathbb{N} \rightarrow \mathbb{R}$ such that for every pattern $\pattern \in \lang$, one has $\uFunction{\measure}(\pattern) = f_\measure(\norm{\pattern})$.
\end{definition}

For instance, the utility \(\uFunction{\mArea}(\pattern) = \norm{\pattern}\) allows to consider the area measure \(\mArea(\pattern, \database) = \mFreq(\pattern, \database) \times \norm{\pattern}\), then one has \(f_{\mArea}(\uLen) =\uLen\). Obviously, the norm-based utility \(\uFunction{\mFreq}(\pattern) = 1\) enables to use the frequency as an interestingness measure. Besides, the utility \(\uFunction{\leq \maxNorm}\) (resp. \(\uFunction{\geq \minNorm}\)) defined as $1$ if \(\norm{\pattern} \leq \maxNorm\) (resp. \(\norm{\pattern} \geq \minNorm\)) and $0$ otherwise, simulates a maximum (resp. minimum) norm constraint. Indeed, with the induced interestingness measure \(\mFreq(\pattern, \database) \times \uFunction{\leq \maxNorm}(\pattern)\) (resp. \(\mFreq(\pattern, \database) \times \uFunction{\geq \minNorm}(\pattern)\)), a pattern with a norm strictly greater than \maxNorm (resp. lower than \minNorm) is judged useless (whatever its frequency). \(\geq \minNorm\) and \(\leq \maxNorm\) are said to be norm-based utility constraints (where 1 means true and 0 means false). The utility \(\mDecay(\pattern) = \alphaDecay^{\norm{\pattern}}\), with $\alphaDecay \in ]0,1]$, named exponential decay, is useful for penalizing long patterns but in a smooth way in comparison with \(\uFunction{\leq \maxNorm}\). Finally, \(\uFunction{\mInvArea}(\pattern) = \frac{1}{\norm{\pattern}}\) allows us to consider the average utility measure.

\noindent {\bf Important remarks:} With weighted items, the utility is not norm-based because the weights $\weight(\literal, \instance)$ of each item $\literal$ depend on the transaction $\instance$ of the database in which it appears. In this case, each pattern $\pattern \subseteq \instance$ has a utility within the transaction $\instance$ defined by $\utility(\pattern, \instance) = \sum_{\literal \in \pattern} \weight(\literal, \instance)$. Therefore, with the language $\langI$, if the items are not weighted, we consider $\utility(\pattern, \instance)=1$ if $\pattern \subseteq \instance$. It is also essential to note that, with sequential data defined over $\langS$, we have $\utility(\pattern, \instance)=1$ if $\pattern \subseteq \instance$ since we do not deal with high utility sequential patterns mining. Obviously, for any pattern language $\lang \in \{\langI, \langS\}$, we consider $\utility(\pattern, \instance) =0$ if $\pattern \not\subseteq \instance$.
Based on these remarks, we introduce the following definition:

\begin{definition}[Norm-based utility measure]
Let \instance be an instance and \pattern a pattern defined over \lang. The norm-based utility measure, also said the interestingness utility measure of \pattern within \(\instance\), denoted by \(\measure(\pattern, \instance)\), is defined as follows: $$\measure(\pattern, \instance) = \utility(\pattern, \instance) \times \uFunction{\measure}(\pattern).$$
\end{definition}

In general, we are interested by the utility of a pattern in the entire database that we call the pattern global utility.

\begin{definition}[Global Pattern Utility]
	\label{def:patGlobUtil}
	Let \database be a database defined over a pattern language \lang and \measure an interestingness utility measure. The global utility of \(\pattern\) in \database is given by:
	\[
	\globalUtilMeasure_\measure(\pattern, \database) = \sum_{(\timestamp, \batch) \in \database} \left( \sum_{\instance \in \batch} \measure(\pattern, \instance)\right).
	\]
\end{definition}

In stream data under temporal biases, the utility of a pattern \pattern inserted at time $\timestamp_j$ can be different to its utility at time $\timestamp_n$, with $n>j$. It depends on what the user really needs to favor, recent or all patterns, by weighting each pattern with a temporal bias. Therefore, we introduce a generic damping function defined as follows:

\begin{definition}[Damping function $\temporalbiasfunction{\dampingFactor}{\timestamp_n, \timestamp_j}$]
    The temporal bias is used when a user need to favor the recent patterns or not. It is based on a damping factor $\dampingFactor \in [0,1]$. At time $\timestamp_n$, the temporal bias of each visited instance $\instance_j$ at time $\timestamp_j \leq \timestamp_n$ are formally updated as follows: $\temporalbiasfunction{\dampingFactor}{\timestamp_n, \timestamp_j}=\expo{-(\timestamp_n-\timestamp_j)\times\dampingFactor}$.
\end{definition}

\noindent We can see that if $\dampingFactor=0$, then $\temporalbiasfunction{\dampingFactor}{\timestamp_n, \timestamp_j}=1$, which corresponds to the landmark window.



To take account these temporal biases in our approach, we define the pattern global utility under temporal bias as follow:

\begin{definition}[Pattern Global Utility under temporal bias]
	\label{def:patGlobUtilunderTemporalBias}
	Let $\database=\sequenceOf{(\timestamp_1, \batch_1), \ldots, (\timestamp_n, \batch_n)}$ be a stream data defined over a pattern language \lang, \measure be an interestingness utility measure and $\dampingFactor \in [0,1]$ a damping factor. At time $\timestamp_n$, the global utility of any pattern \pattern inserted into the reservoir at time $\timestamp$ and that still appears into \reservoir is given by:
	\[
	\globalUtilMeasure_\measure^{\dampingFactor}(\pattern, \database) = \sum_{(\timestamp_i, \batch_i) \in \database}\left( \left( \sum_{\instance_{i_j} \in \batch_i} \measure(\pattern, \instance_{i_j})\right) \times \temporalbiasfunction{\dampingFactor}{\timestamp_n,\timestamp_i}\right).
	\]
\end{definition}

\begin{example}
    Table \ref{tab:toy_seq} and Table \ref{tab:toy_wItemset} present two toy datasets respectively for sequential and weighted itemsets. They also give some measures with temporal biases and utility measures. The damping factor is set to $\dampingFactor \in \{0, 0.1\}$. For Table \ref{tab:toy_wItemset}, $\{A, B, C\}$, $\{2, 1.5, 2\}$ means that the items $A$, $B$, and $C$ have weights $2$, $1.5$ and $2$ respectively in the instance $\instance_1$.
    \begin{table}[htp!]
    {
	\centering
	\caption{Stream batches of sequential itemsets}
	\label{tab:toy_seq}
	\begin{tabular}{c|c|c|c}
	$\downarrow$ time & id & Sequential itemsets & $\utility(\sequenceOf{\{A\}\{C\}}, \instance_i)$\\
	\hline
	\multirow{2}{*}{$(\timestamp_1, \batch_1)$} & $\instance_1$ & $\sequenceOf{\{\underline{A}\}\{B\}\{A,\underline{C}\}\{B\}}$ & $1$ \\
		& $\instance_2$ & $\sequenceOf{\{\underline{A}, B, C\}\{\underline{C}\}\{A,C\}}$ & $1$ \\
	\hline
	$(\timestamp_2, \batch_2)$	& $\instance_3$ & $\sequenceOf{\{B\}\{A,C\}\{A\}}$ & $0$\\
	\hline\hline
    \multicolumn{2}{|l|}{$\globalUtilMeasure_{\mFreq}(\sequenceOf{\{A\}\{C\}}, \database)$} & \multicolumn{2}{r|}{$1 + 1 + 0 = 2$}\\
    \multicolumn{2}{|l|}{$\globalUtilMeasure_{\mArea}^{0}(\sequenceOf{\{A\}\{C\}}, \database)$} & \multicolumn{2}{r|}{$(1\times 2 + 1\times 2) \times 1  + (0 \times 2)\times 1 = 4$}\\
    \multicolumn{2}{|l|}{$\globalUtilMeasure_{\mFreq}^{0.1}(\sequenceOf{\{A\}\{C\}}, \database)$} & \multicolumn{2}{r|}{$(4) \cdot \expo{-(2-1)\cdot 0.1} + (0)\cdot \expo{-(2-2)\cdot 0.1}\approx 3.6$}\\
    \hline
	\end{tabular}

    \smallskip

    \smallskip
 
	\caption{Stream batches of weighted itemsets}
	\label{tab:toy_wItemset}
	\begin{tabular}{c|c|c|c}
	$\downarrow$ time & id & Weighted itemsets & $\utility(\{B, C\}, \instance_i)$ \\
	\hline
	$(\timestamp_1, \batch_1)$	& $\instance_1$ & $\{A, B, C\}$, $\{2, 1.5, 2\}$ & $1.5+2=3.5$ \\
	\hline
	\multirow{2}{*}{$(\timestamp_2, \batch_2)$}& $\instance_2$ & $\{A, C\}$, $\{3, 3\}$  & $0$ \\
		& $\instance_3$ & $\{B, C, D, E\}$, $\{2, 1, 2, 1\}$ & $2+1=3$ \\
	\hline\hline
    \multicolumn{2}{|l|}{$\globalUtilMeasure_{\mUtility}(\{B,C\}, \database)$} & \multicolumn{2}{r|}{$3.5 + 0 + 3 = 6.5$}\\
    \multicolumn{2}{|l|}{$\globalUtilMeasure_{\mAvgUtility}^{0}(\{B, C\}, \database)$} & \multicolumn{2}{r|}{$(3.5\times \frac{1}{2} + 0\times \frac{1}{2}) \times 1  + (3 \times \frac{1}{2})\times 1 = \frac{6.5}{2}$} \\
    \multicolumn{2}{|l|}{$\globalUtilMeasure_{\mUtility}^{0.1}(\{B, C\}, \database)$} & \multicolumn{2}{r|}{$(3.5) \cdot \expo{-(2-1)\cdot 0.1}  + (3)\cdot \expo{-(2-2)\cdot 0.1} \approx 6.2$}  \\
    \hline
	\end{tabular}
    }
\end{table}
\end{example}

\subsection{Reservoir-based multi-step pattern sampling problem}
A reservoir-based pattern sampling aims to randomly maintains a sample of patterns proportionally to a given utility measure. We provide in Table \ref{tab:summary} some of the most interesting utility measures presented in the literature \cite{BLPG11,diop2019kais,DiopPAKDD2022}. Here, $\instance$ represents an instance of the language $\lang$, and $\pattern$ is a pattern in the same language. For instance, in the case of area measure without constraints, we have $\mArea(\pattern, \instance) = \norm{\pattern}$, and the global utility under landmark temporal bias of $\pattern$ in $\database$ is given by $\globalUtilMeasure_\measure^{0}(\pattern, \database)=\sum_{(\timestamp, \batch) \in \database}(\sum_{\instance \in \batch}\mArea(\pattern, \instance))\times \temporalbiasfunction{0}{\timestamp_n, \timestamp} = \mFreq(\pattern, \database) \times \norm{\pattern}$. This measure is applicable to both types of pattern languages, $\langI$ and $\langS$. Previous research has shown that combining several norm-based utility measures results in another norm-based utility measure \cite{diop2019kais}. Therefore, when imposing norm constraints (minimal and/or maximal) to tackle the long-tail problem \cite{diopICDM18}, norm-based utility measures are a suitable choice. For example, $\measure = (\mArea) \times (\mMinNorm) \times (\mMaxNorm)$ does not allow a pattern $\pattern$ for which $\norm{\pattern} \not\in [\minLen..\maxLen]$, and such patterns have a utility of $\globalUtilMeasure_\measure^{\dampingFactor}(\pattern, \database)=0$.

\begin{table*}
\begin{center}
    
	\caption{Definition of some useful interestingness utility measures}
	\label{tab:summary}
	\begin{tabular}{|l|c|c|c|c|c|}
		\hline
	Measure (\measure)	& $\utilityTrans(\pattern, \instance)$  & $\uFunction{\measure}(\pattern)$ & $\globalUtilMeasure_\measure(\pattern, \database)$ & Language \lang & Under interval norm constraint\\
		\hline
		\hline
	Frequency (\mFreq)\cite{Agrawal93}	& 1 & 1 & $\mFreq(\pattern, \database)$ & \langI, \langS & $\measure = (\mFreq) \times (\mMinNorm) \times (\mMaxNorm)$ \\
	\hline
	Area (\mArea) \cite{Geerts2004TilingD}	& 1 & \norm{\pattern} & $\mFreq(\pattern, \database) \times \norm{\pattern}$ & \langI, \langS & $\measure = (\mArea) \times (\mMinNorm) \times (\mMaxNorm)$ \\
	\hline
	Exponential Decay (\mDecay) \cite{NakagawaExpDecay}	& 1 & $\alphaDecay^{\norm{\pattern}}$ with $\alphaDecay \in ]0,1]$ & $\mFreq(\pattern, \database) \times \alphaDecay^{\norm{\pattern}}$ & \langI, \langS & $\measure = (\mDecay) \times (\mMinNorm) \times (\mMaxNorm)$ \\
	\hline
	High Utility (\mUtility) \cite{RaymondHUI} & $\utility(\pattern, \instance)$ & 1 & $\sum_{(\timestamp, \instance) \in \database} \utility(\pattern, \instance)$ & \langI & $\measure = (\mUtility) \times (\mMinNorm) \times (\mMaxNorm)$ \\
	\hline
	High Average Utility (\mAvgUtility)\cite{TruongHAUI}	& $\utility(\pattern, \instance)$ & $\frac{1}{\norm{\pattern}}$ & $\sum_{(\timestamp, \instance) \in \database} \frac{\utility(\pattern, \instance)}{\norm{\pattern}}$ & \langI & $\measure = (\mAvgUtility) \times (\mMinNorm) \times (\mMaxNorm)$ \\
		\hline
	\end{tabular}
\end{center}
\end{table*}

We focus on a generic solution that can handle reservoir-based sequential patterns and high (average) utility patterns sampling with a multi-step random procedure under norm-based utility measures and temporal biases. 
The problem addressed in this paper can be formally stated as follows:

\begin{problem}[Reservoir pattern sampling]
{ Given a stream of data $\database = \sequenceOf{(\timestamp_1, \batch_1), \ldots, (\timestamp_n, \batch_n)}$ defined over a pattern language $\lang \in \{\langI, \langS\}$, a damping factor $\dampingFactor \in [0,1]$, and an interestingness utility measure \measure, our goal is to maintain a reservoir of size \reservoirSize patterns $[\pattern_1, \ldots, \pattern_\reservoirSize]$ from \lang. Each pattern $\pattern_j$ should be drawn with a probability proportional to its weighted utility in the database \database, defined as:
	$$\prob(\pattern_j, \database)= \frac{\globalUtilMeasure_\measure^{\dampingFactor}(\pattern_j, \database)}{\sum_{\pattern \in \lang(\database)} \globalUtilMeasure_\measure^{\dampingFactor}(\pattern, \database)}.$$
}
 \end{problem}
\section{Reservoir-based Multi-Step Pattern Sampling}
\label{sec:contrib}
A multi-step pattern sampling approach is a sampling technique with replacement. Therefore, in this paper, we focus on weighted reservoir sampling with replacement.

\subsection{A breakdown of the three steps of our approach}

\begin{figure}
	\centering
 
	\includegraphics[width=0.33\textwidth, keepaspectratio]{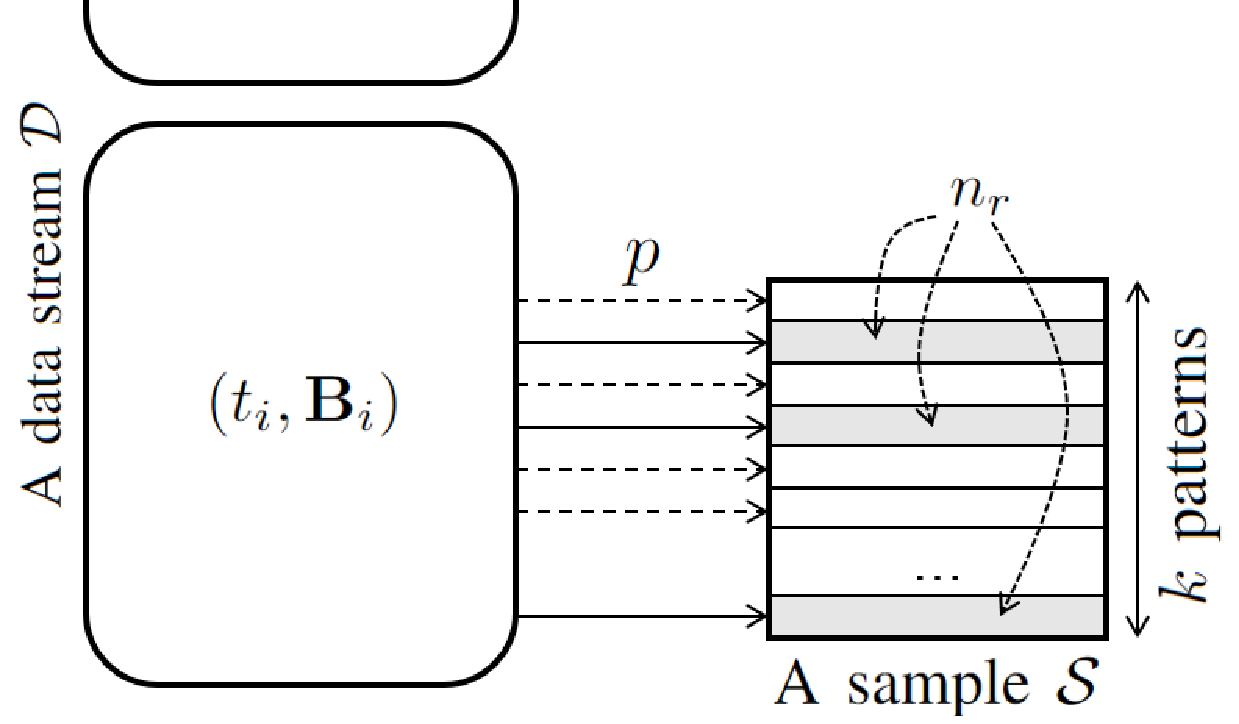}
 \caption{Overview of the approach \textit{(the incomplete block denotes next batch)}}
\label{fig:overview}
\end{figure}

Weighted reservoir sampling \cite{efraimidis2006weighted} has been proposed to maintain into a reservoir with fixed size a sample of weighted data points where each data point is maintained with a probability proportional to its weight. 
For this purpose, Figure~\ref{fig:overview} depicts the overview of the approach. Intuitively, when a new batch $\batch_{i}$ arrives from the data stream at time $t_i$, the method calculates the probability $p$ that a pattern in the reservoir \reservoir will be replaced by a new pattern drawn from $\batch_{i}$ (Step 1). Interestingly, rather than going through each of the $k$ patterns in \reservoir one by one, the number of patterns to be replaced $n_r$ can be determined directly (Step 2). Then, $n_r$ patterns are drawn from the batch $\batch_{i}$ using a traditional pattern sampling method to randomly replace $n_r$ patterns from the reservoir (Step 3).
We detail below the three steps that our approach should follows.

\smallskip

\paragraph{{\bf Step 1.} Batch acceptance probability} Let us assume that all the positions of the reservoir are already occupied by a pattern from past batches $\sequenceOf{(\timestamp_1, \batch_1), \ldots, (\timestamp_n, \batch_{j-1})})$ and that all batch weights $\weight_\measure(\batch_i) = \sum_{\instance_{i_{j'}} \in \batch_i}\sum_{\pattern \specialization \instance_{i_{j'}}}\measure(\pattern, \instance_{i_{j'}})$, with $i \in [1..j-1]$ are feasible. For any batch $\batch_j$ with a weight $\weight_\measure(\batch_j)$, we compute the probability acceptance that of one of its patterns from $\lang(\batch_j)$ be inserted into the reservoir $\acceptanceProb_j$. Based on \cite{efraimidis2006weighted}, we have $\acceptanceProb_j = \frac{\weight_\measure(\batch_j)}{\sum_{i=1}^{j} \weight_\measure(\batch_i)\times \temporalbiasfunction{\dampingFactor}{\timestamp_j, \timestamp_i}}$. Therefore, a pattern drawn proportionally to its weight in $\batch_j$ can be inserted at a position of the reservoir uniformly drawn. This uniform replacement has been already used by A-Chao \cite{chao1982reservoir} for weighted reservoir sampling in data stream.

However, it is evident that computing the acceptance probability by iteratively summing the weights of the visited batches is infeasible because we do not store the past batches. Therefore, we employ a memory-less computing technique to adapt the normalization constant that avoids storing information for each batch received.

\begin{property}
	\label{prop:dampedWindow} 
	Let $\normalization_{i-1}$ be the normalization constant for the $i-1$ first batches of the data stream under the damped window and the norm-based utility measure \measure, with  $\normalization_{0} = 0$. The probability $p$ to draw a pattern from the next batch $\batch_{i}$ under the damped window can be computed as follows:
	
	$p=\frac{\weight_\measure(\batch_{i}) \times \expo{\dampingFactor \times \timestamp_{i}}}{\normalization_{i}}$, with $\normalization_{i} = \normalization_{i-1} + \weight_\measure(\batch_{i})\times \expo{\dampingFactor \times \timestamp_{i}}$.
\end{property}

\begin{proof} We omit the proof due to space constraints.
\end{proof}


    

After the acceptance of batch $\batch_j$ with a probability of $p_j$ in Step 1, we know that at least one of its pattern set should be inserted into the reservoir. But, there is also the possibility of other patterns from the set being inserted at various positions within the reservoir. This drives the rationale behind the second step of our approach.

\smallskip

\paragraph{{\bf Step 2.} Number of patterns to draw from an accepted batch} Let \reservoir be a reservoir of size \reservoirSize where we need to store a sample from a population of finite size. In this case, each pattern can be selected up to $\numberOfRealisations \leq \reservoirSize$ times in the sample. To achieve this goal, we process \reservoirSize copies of $\lang(\batch_j)$ such in each of them, a pattern is drawn proportionally to its weight in $\batch_j$. Interestingly, using \reservoirSize copies of $\lang(\batch_j)$ with a probability $p_j$ corresponds to simulating \reservoirSize independents Bernoulli trials with a probability $p_j$. By definition, the probability of obtaining \numberOfRealisations success trials is nothing else that the Binomial Distribution which is formalized as follows:

$$
	\prob(X=\numberOfRealisations)= \binom{\reservoirSize}{\numberOfRealisations} p_j^\numberOfRealisations (1 - p_j)^{\reservoirSize-\numberOfRealisations}
$$

with \numberOfRealisations is the number of successful trials (or the number of patterns selected for inclusion in the reservoir), \reservoirSize the size of the reservoir (the total number of trials or positions in the reservoir), and $p_j=\frac{\weight_\measure(\batch_j)}{\sum_{i=1}^{j} \weight_\measure(\batch_i)}$ the probability of success in each trial, which is the probability of a pattern of $\lang(\batch_j)$ being selected for inclusion in the reservoir. 
However, we note that computing a probability acceptance for each position is time consuming. Therefore, to skip computing an acceptance rate for each position, we use the Cumulative Binomial Probability Distribution defined as follows: 

\begin{definition}[Cumulative Binomial Probability Distribution (CBPD)]
	Suppose there is an event with a probability \(p\) of occurring per trial. The cumulative binomial probability \prob, representing the probability of this event occurring \numberOfRealisations or more times in \reservoirSize trials, is as follows:
	\begin{equation}
		\label{eq:binom}
		\prob(\numberOfRealisations, \reservoirSize) \equiv \sum_{i=\numberOfRealisations}^{\reservoirSize} \binom{\reservoirSize}{i} p^i (1 - p)^{\reservoirSize-i}.
	\end{equation}
\end{definition}

This formula calculates the cumulative probability of obtaining \numberOfRealisations or more successful trials out of \reservoirSize trials. It does so by summing the probabilities of all possible outcomes from \numberOfRealisations to \reservoirSize, where each outcome represents a different number of successful trials. In this case, $p^i$ represents the probability of having $i$ successes, and $(1 - p)^{\reservoirSize-i}$ represents the probability of having $\reservoirSize -i$ failures.

In advance, the CBPD  can be more efficiently computed by using the incomplete beta function (IBF), $\incBetaFunc{a}{b}{x}$\cite{AbramowitzStegun1972}. The IBF represents the probability that a random variable following a beta distribution with parameters $a$ and $b$ falls below the value $x$. By leveraging the IBF, we can efficiently handle complex probability calculations without explicit summation of individual probabilities (as done in Eq. \ref{eq:binom}), minimizing computational load and numerical errors. 

\begin{definition}[Incomplete beta function (IBF) \cite{Didonato1967TheEC}]
	The incomplete beta function $I_x(a, b)$ is defined as follows:
	\[
	\incBetaFunc{a}{b}{x} = \frac{1}{B(a, b)} \int_{0}^{x} t^{a-1}(1-t)^{b-1} \, dt
	\]
	
	\noindent where $a$ and $b$ are positive real numbers (parameters of the beta distribution); $B(a, b)$ is the beta function, defined as: $$B(a, b) = \int_{0}^{1} t^{a-1}(1-t)^{b-1} \, dt $$ and $x$ is a real number in the range $[0,1]$.
\end{definition}


\begin{property}[From CBPD to IBF]
	\label{prop:CBPD-IBF}
	Let \reservoirSize be the total number of trials, $\numberOfRealisations \leq \reservoirSize$ the minimum number of times the event must occur, and $p$ the probability of the event occurring in a single trial. The Cumulative Binomial Probability Distribution can be computed as follows:
	\begin{equation}
		\label{eq:IBF}
		\prob(\numberOfRealisations, \reservoirSize) \equiv \sum_{j=\numberOfRealisations}^{\reservoirSize} \binom{\reservoirSize}{j} p^j (1 - p)^{\reservoirSize-j} = \incBetaFunc{\numberOfRealisations}{\reservoirSize - \numberOfRealisations + 1}{p}.
	\end{equation}
\end{property}

\begin{proof} We omit the proof due to space constraints.
\end{proof}

\begin{definition}[Inverse Incomplete Beta Function]
\label{def:inverse_ibf}
The Inverse Incomplete Beta Function allows for the approximation of the number of successful trials $\numberOfRealisations$ out of $\reservoirSize$ trials that matches the CBPD with a given probability $x \in [0,1]$ as follows: 
$$\numberOfRealisations = \arg_{\numberOfRealisations}[\incBetaFunc{\numberOfRealisations}{\reservoirSize - \numberOfRealisations + 1}{\acceptanceProb}=x].$$
\end{definition}



Now we are going to show how to draw a pattern proportionally to its interest from an accepted batch.

\paragraph{{\bf Step 3.} First pattern occurrence sampling from a batch} The main goal of processing a copy of an acceptance batch is finally to draw a pattern. However, the complexity of to draw a pattern from a batch depends to the pattern language. With sequential itemsets, a pattern can have multiple occurrences within a sequence \cite{diopICDM18} which is not the case with weighted/unweighted itemsets. Since we propose a generic approach dealing with sequential itemsets, then we adapt the first occurrence definition previously introduced in \cite{diopICDM18}.

\begin{definition}[First occurrence]
\label{def:first_occ}
Given an instance $\instance$, let $o_1$ and $o_2$ be two occurrences of a pattern $\pattern$ within $\instance$, whose signatures are $\langle i^1_1, i^1_2, \ldots, i^1_N \rangle$ and $\langle i^2_1, i^2_2, \ldots, i^2_N \rangle$ respectively. $o_1$ is less than $o_2$, denoted by $o_1 < o_2$, if there exists an index $k \in [1..N]$ such that for all $j \in [1..k - 1]$, one has $i^1_j = i^2_j$, and $i^1_k < i^2_k$. Finally, the first occurrence of $\pattern$ in $\instance$ its smallest occurrence with respect to the order defined previously.
\end{definition}

\begin{example}
For instance, $\pattern=\sequenceOf{\{A\}\{C\}}$ has two occurrences  $o_1$ and $o_2$ in $\instance_2=\sequenceOf{\{{A}, B, C\}\{{C}\}\{A,C\}}$ with signatures \sequenceOf{1,2} and \sequenceOf{1,3} respectively. But $o_1$ is the first occurrence because $o_1 < o_2$ since $2<3$.
\end{example}

Based on Definition \ref{def:first_occ}, we then propose a generic sampler operator named $\sample{\lang}{\batch}{\measure}{\numberOfRealisations}$. In fact, run $\numberOfRealisations$  times the operator $\sample{\lang}{\batch}{\measure}{1}$ in the same batch \batch in order to get a first pattern occurrence for each realisation is equivalent to run $\sample{\lang}{\batch}{\measure}{\numberOfRealisations}$ because $\bigcup_{i=1}^{\numberOfRealisations}\{\pattern_i \sim \measure(\lang, \batch)\} \equiv \sample{\lang}{\batch}{\measure}{\numberOfRealisations}$. Algorithm \ref{alg:op-sample} implements the operator $\sample{\lang}{\instance}{\measure}{\numberOfRealisations}$, which is used to draw \numberOfRealisations patterns from the set of patterns in \batch, with each pattern drawn proportionally to the interestingness measure \measure. In line 1, each instance is weighted by the sum of its patterns utility. First, an instance \instance is drawn proportionally to its weight in line 4. Then, in line 5, each norm \lengthOrNorm is weighted based on the sum of pattern utilities of norm \lengthOrNorm in \instance. To draw a pattern from \instance, an integer $\lengthOrNorm'$ is first selected proportionally to its weight in \instance (line 6). At line 7, a first occurrence of a pattern of norm $\lengthOrNorm'$ is drawn proportionally to the set of patterns of norm $\lengthOrNorm'$ in \instance, i.e., $\prob(\pattern_j|\instance, \lengthOrNorm')=\frac{\measure(\pattern_j, \instance)}{\weight_\measure^\instance(\lengthOrNorm')}$, and added to the sample (line 8). This process (lines 4-8) is repeated \numberOfRealisations times. Finally, a sample of \numberOfRealisations patterns is returned at line 9.

{\small
\begin{algorithm}
	\caption{$\sample{\lang}{\batch}{\measure}{\numberOfRealisations}$}
	\label{alg:op-sample}
	\begin{algorithmic}[1]
		\Statex {\bf Output:}  A sample of \numberOfRealisations patterns of $\lang(\batch)$ drawn proportionally to the utility measure \measure
		\State Let $\weight_\measure(\instance) \gets \sum_{\pattern \specialization \instance} \measure(\pattern,\instance)$ for all $\instance \in \batch$
		\State Let $\phi \gets \emptyset$
		\For{$j \in [1..\numberOfRealisations]$}
		\State Draw \instance from \batch with $\prob(\instance,\batch)=\frac{\weight_\measure(\instance)}{\sum_{\instance_i \in \batch}\weight_\measure(\instance_i)}$
		\State Let $\weight_\measure^\instance(\lengthOrNorm) \gets \sum_{\pattern \specialization \instance \wedge \norm{\pattern}=\lengthOrNorm} \measure(\pattern,\instance)$ for $\lengthOrNorm \in [1..\norm{\instance}]$ 
		\State Draw an integer $\lengthOrNorm'$ with a probability of $\frac{\weight_\measure^\instance(\lengthOrNorm')}{\sum_{\lengthOrNorm}\weight_\measure^\instance(\lengthOrNorm)}$
		\State Let $\pattern_j \sim \measure(\{\pattern\specialization \instance: \norm{\pattern}=\lengthOrNorm'\})$
		\State $\phi \gets \phi \cup \{\pattern_j\}$
		\EndFor
		\State  \Return{$\phi$}
	\end{algorithmic}
\end{algorithm}
}

We are now going to present our generic algorithm based on a weighted reservoir sampling with replacement.

\subsection{A Generic {R}eservoir-based {T}hree-Step Pattern {S}ampling}
\label{sub:myalgo}

We first give a high level description of \myalgo described in Algorithm \ref{alg:spasx}. It takes a data stream \database, a utility measure \measure, the desired reservoir size \reservoirSize, and a damping factor $\dampingFactor \in [0,1]$. First, for each batch $\batch_i$ appearing at timestamp $\timestamp_i$, the acceptance probability $\acceptanceProb$ that a pattern from $\lang(\batch_i)$ replaces a pattern inserted at $\timestamp_{i'}$, with $\timestamp_{i'} < \timestamp_{i}$ is computed (lines 3-5). 
If $\batch_i$ is accepted (line 6), which correspond to a success trial, then the number of additional success trials out of the rest of the reservoir size $\reservoirSize - 1$ that a pattern of $\lang(\batch_i)$ should be inserted (line 7) is deduced based on the inverse IBF (Definition \ref{def:inverse_ibf}). 
Lines 8 to 11 allow to draw \numberOfRealisations patterns with replacement where each draw corresponds to an inserted pattern. At time $\timestamp_n$, \myalgo maintains a reservoir of \reservoirSize patterns where each pattern \pattern is selected with a probability proportional $\globalUtilMeasure_\measure^{\dampingFactor}(\pattern, \database)$.

\begin{algorithm}
	\caption{\myalgo: A Generic Stream pattern sampler}
	\label{alg:spasx}
	\begin{algorithmic}[1]
		\Statex {\bf Input:}  A data stream \database, a utility measure \measure, a damping factor $\dampingFactor \in [0,1]$, and the desired reservoir size \reservoirSize 
		
		\Statex {\bf Output at time $\timestamp_n$:}  A sample \reservoir of \reservoirSize patterns drawn in $\lang(\database=\sequenceOf{(\timestamp_1, \batch_1), \ldots, (\timestamp_n, \batch_n)})$ based on \measure and \dampingFactor
		
		\State $\reservoir \gets \emptyset$; $\normalization_{0}=0$
		\While {$(\timestamp_i, \batch_i)$ is from \database} 
		\Statex {~~~~//{\bf Batch acceptance probability}}
		\State 
       $\normalization_{i} = \normalization_{i-1} + \weight_\measure(\batch_{i})\times \expo{\dampingFactor \times \timestamp_{i}}$
		\State $\acceptanceProb \gets \frac{\weight_\measure(\batch_i) \times \expo{\dampingFactor \times \timestamp_{i}}}{\normalization_{i}}$
		\State $x \gets random(0,1)$
		\If{$p> x$}
		\Statex {~~~~~~~~//{\bf Number of realisations}}
		\State $\numberOfRealisations \gets 1+ \underset{\numberOfRealisations}{\arg} [\incBetaFunc{\numberOfRealisations}{\reservoirSize - \numberOfRealisations}{\acceptanceProb}=x]$ \Comment{Definition \ref{def:inverse_ibf}} 
		\Statex {~~~~~~~~//{\bf Patterns selection}}
		\State $ E \gets getPatternsToRemove(\reservoir, \numberOfRealisations)$
		\State $\reservoir \gets \reservoir \setminus \{\reservoir[j]: \text{ for } j \in E\}$
        \For{$\pattern_{j} \in \sample{\lang}{\batch_i}{\measure}{\numberOfRealisations}$} 
		\State $\reservoir \gets \reservoir \cup \{(\timestamp_i, \pattern_{j}) \}$
		\EndFor
		\EndIf
		\EndWhile
	\end{algorithmic}
\end{algorithm}

Regarding the $getPatternsToRemove$, it returns \numberOfRealisations distinct indexes uniformly drawn from $[1..\card{\reservoir}]$. Because, thanks to \cite{Efraimidis2015}, all patterns in the reservoir have an equal probability of being replaced by one of the \numberOfRealisations patterns drawn from the current batch by the sampler $\sample{\lang}{\batch_i}{\measure}{\numberOfRealisations}$.

\subsection{Theoretical analysis of the method}
\label{sub:analysis}
We study now the soundness and the complexity of \myalgo.
\paragraph{{Soundness analysis}}
The following properties state that \myalgo returns an exact
sample of patterns under temporal bias with norm-based utility measure.

Let us first demonstrate that the batch acceptance probability computation of Algorithm \ref{alg:spasx} is exact.

\begin{property}[Batch acceptance probability]
	\label{prop:Step1Soundness}
	Given a stream data \database defined over a pattern language \lang, $\database=\sequenceOf{(\timestamp_1, \batch_1), \ldots, (\timestamp_n, \batch_n)}$, \measure be a norm-based utility measure, $\dampingFactor \in [0,1]$ a damping factor, and \reservoirSize the size of the reservoir \reservoir. After observing $(\timestamp_n, \batch_n)$, the probability that a pattern of batch $\batch_{i}$, inserted at time $\timestamp_i$ at the $\text{j}^{\text{th}}$ position of the reservoir, $j \in [1..\reservoirSize]$, stays in \reservoir, denoted $\prob(\batch_i \insertPattern \reservoir[j]| \timestamp_n)$, is given by: $\prob(\batch_i \insertPattern \reservoir[j]| \timestamp_n)=\frac{\weight_\measure(\batch_i)\times \temporalbiasfunction{\dampingFactor}{\timestamp_n,\timestamp_i} }{\sum_{i'\leq n}\weight_\measure(\batch_{i'})\times \temporalbiasfunction{\dampingFactor}{\timestamp_n,\timestamp_{i'}}}$.
\end{property}

\begin{sloppypar}
	
\begin{proof}
	We know that $\prob(\batch_i \insertPattern \reservoir[j]| \timestamp_n)= \prod_{i'=i}^n\prob(\batch_i \insertPattern \reservoir[j]| \timestamp_{i'})$ and $\temporalbiasfunction{\dampingFactor}{\timestamp_n,\timestamp_i}=\expo{-(\timestamp_n - \timestamp_i) \times \dampingFactor}$, with $\dampingFactor \in [0,1]$. Thanks to Property \ref{prop:dampedWindow}, we have $\prob(\batch_i \insertPattern \reservoir[j]| \timestamp_n)=\frac{\weight_\measure(\batch_i)\times \expo{\timestamp_i\times \dampingFactor}}{\normalization_{i}} \times \frac{\normalization_{i+1} - \weight_\measure(\batch_{i+1})\times \expo{\timestamp_{i+1}\times \dampingFactor}}{\normalization_{i+1}} \times \ldots \times \frac{\normalization_{n-1} - \weight_\measure(\batch_{n-1})\times \expo{\timestamp_{n-1}\times \dampingFactor}}{\normalization_{n-1}}\times \frac{\normalization_{n} - \weight_\measure(\batch_n)\times \expo{\timestamp_{n}\times \dampingFactor}}{\normalization_{n}}$. Based on the same property, we also have $\normalization_{i+1} - \weight_\measure(\batch_{i+1})\times \expo{\timestamp_{i+1}\times \dampingFactor} = \normalization_{i}$. Then,
	$\prob(\batch_i \insertPattern \reservoir[j] | \timestamp_n)=\frac{\weight_\measure(\batch_i)\times \expo{\timestamp_i\times \dampingFactor}}{\normalization_{i}} \times \frac{\normalization_{i}}{\normalization_{i+1}} \times \ldots \times \frac{\normalization_{n-2} }{\normalization_{n-1}}\times \frac{\normalization_{n-1}}{\normalization_{n}}=\frac{\weight_\measure(\batch_i)\times \expo{\timestamp_i\times \dampingFactor}}{\normalization_{n}}$. We know that $\frac{\normalization_{n}}{\expo{\timestamp_n \times \dampingFactor}} = \sum_{i'\leq n}\weight_\measure(\batch_{i'})\times \temporalbiasfunction{\dampingFactor}{\timestamp_n,\timestamp_{i'}}$. Then, $\prob(\batch_i \insertPattern \reservoir[j] | \timestamp_n)=\frac{\weight_\measure(\batch_i)\times \expo{\timestamp_i\times \dampingFactor}}{\expo{\timestamp_n \times \dampingFactor}\times \sum_{i'\leq n}\weight_\measure(\batch_{i'})\times \temporalbiasfunction{\dampingFactor}{\timestamp_n,\timestamp_{i'}}} = \frac{\weight_\measure(\batch_i)\times \expo{\timestamp_i\times \dampingFactor} \times \expo{-\timestamp_n \times \dampingFactor}}{\sum_{i'\leq n}\weight_\measure(\batch_{i'})\times \temporalbiasfunction{\dampingFactor}{\timestamp_n,\timestamp_{i'}}} = \frac{\weight_\measure(\batch_i)\times \expo{-(\timestamp_n - \timestamp_i)\times \dampingFactor}}{\sum_{i'\leq n}\weight_\measure(\batch_{i'})\times \temporalbiasfunction{\dampingFactor}{\timestamp_n,\timestamp_{i'}}} = \frac{\weight_\measure(\batch_i)\times \temporalbiasfunction{\dampingFactor}{\timestamp_n,\timestamp_i}}{\sum_{i'\leq n}\weight_\measure(\batch_{i'})\times \temporalbiasfunction{\dampingFactor}{\timestamp_n,\timestamp_{i'}}}$. According to these two cases, we can conclude that Property \ref{prop:Step1Soundness} holds for landmark window and damped window and for any norm-based utility measure.
\end{proof}

\end{sloppypar}

Let us now show that the operator $\sample{\lang}{\batch}{\measure}{1}$ used in the pattern selection is also exact.

\begin{property}[Pattern selection]
\label{prop:step_3}
	Given a batch \batch defined over a pattern language \lang, and an interestingness utility measure \measure, the operator $\sample{\lang}{\batch}{\measure}{1}$ draw a pattern proportionally to its utility in \batch:  $\prob(\pattern|\batch)=\frac{\measure(\pattern,\batch)}{\sum_{\instance' \in \batch}\weight_\measure(\instance')}$, with $\measure(\pattern,\batch) =\sum_{\instance \in \batch} \measure(\pattern, \instance))$.
\end{property}

\begin{proof}
	By definition, one has $\prob(\pattern|\batch) = \sum_{(\timestamp, \instance) \in \batch} \prob(\instance|\batch) \times \prob(\lengthOrNorm'|\instance) \times \prob(\pattern|\instance, \lengthOrNorm')$. We know that $\prob(\instance|\batch)=\frac{\weight_\measure(\instance)}{\sum_{\instance' \in \batch}\weight_\measure(\instance')}$. Based on the norm-based utility, it can be decomposed as $\prob(\instance|\batch)=\frac{\sum_{\lengthOrNorm}\weight_\measure^\instance(\lengthOrNorm)}{\sum_{\instance' \in \batch}\weight_\measure(\instance')}$. Since $\prob(\lengthOrNorm'|\instance) = \frac{\weight_\measure^\instance(\lengthOrNorm')}{\sum_{\lengthOrNorm}\weight_\measure^\instance(\lengthOrNorm)}$ and $\prob(\pattern|\instance, \lengthOrNorm') = \frac{\measure(\pattern, \instance)}{\weight_\measure^\instance(\lengthOrNorm')}$. Therefore, we have $\prob(\pattern|\batch)= \sum_{\instance \in \batch} \frac{\sum_{\lengthOrNorm}\weight_\measure^\instance(\lengthOrNorm)}{\sum_{\instance' \in \batch}\weight_\measure(\instance')} \times \frac{\weight_\measure^\instance(\lengthOrNorm')}{\sum_{\lengthOrNorm}\weight_\measure^\instance(\lengthOrNorm)} \times \frac{\measure(\pattern, \instance)}{\weight_\measure^\instance(\lengthOrNorm')}$ which leads to $ \prob(\pattern|\batch)=\sum_{\instance \in \batch} \frac{\measure(\pattern, \instance)}{\sum_{\instance' \in \batch}\weight_\measure(\instance')} = \frac{ \sum_{\instance \in \batch} \measure(\pattern, \instance)}{\sum_{\instance' \in \batch}\weight_\measure(\instance')} =\frac{\measure(\pattern,\batch)}{\sum_{\pattern \specialization \instance} \measure(\pattern,\instance)}$. Hence the result.
\end{proof}

Based on these properties, we can proof the soundness of our main algorithm.

\begin{property}[Soundness]
	\label{prop:Soundness}
	Let $\database=\sequenceOf{(\timestamp_1, \batch_1), \ldots, (\timestamp_n, \batch_n)}$ be a stream data defined over a pattern language \lang, \measure be a norm-based utility measure, \reservoirSize the size of the reservoir \reservoir and $\dampingFactor \in [0,1]$ a damping factor. After observing $(\timestamp_n, \batch_n)$, \myalgo returns a sample of \reservoirSize patterns $\pattern_1, \ldots, \pattern_\reservoirSize$ where each pattern $\pattern_j=\reservoir[j]$ is drawn with a probability equals to:  $$\prob(\pattern_j=\reservoir[j], \database)= \frac{\globalUtilMeasure_\measure^{\dampingFactor}(\pattern_j, \database)}{\sum_{\pattern \in \lang} \globalUtilMeasure_\measure^{\dampingFactor}(\pattern_j, \database)}.$$
\end{property}

\begin{sloppypar}
\begin{proof}
	We know that $\prob(\pattern_j=\reservoir[j], \database)= \sum_{(\timestamp_i, \batch_i) \in \database} \prob(\batch_i \insertPattern \reservoir[j] | \timestamp_n) \times \prob(\pattern_j|\batch_i)$, with $\prob(\pattern_j|\batch_i)$ the probability to draw $\pattern_j$ from $\batch_i$ proportionally to the norm-based utility measure \measure given in Property \ref{prop:step_3}, i.e. $\prob(\pattern_j|\batch_i) = \frac{\measure(\pattern_j, \batch_i)}{\weight_\measure(\batch_i)}$ (line 4 of Algorithm \ref{alg:op-sample}). According to Property \ref{prop:Step1Soundness}, we have $\prob(\batch_i\insertPattern \reservoir[j] | \timestamp_n) =\frac{\weight_\measure(\batch_i)\times \temporalbiasfunction{\dampingFactor}{\timestamp_n,\timestamp_i} }{\sum_{j\leq n}\weight_\measure(\batch_j)\times \temporalbiasfunction{\dampingFactor}{\timestamp_n,\timestamp_j}}$. Therefore, we have $\prob(\pattern_j=\reservoir[j], \database)=\sum_{(\timestamp_i, \batch_i) \in \database} \frac{\weight_\measure(\batch_i)\times \temporalbiasfunction{\dampingFactor}{\timestamp_n,\timestamp_i} }{\sum_{j\leq n}\weight_\measure(\batch_j)\times \temporalbiasfunction{\dampingFactor}{\timestamp_n,\timestamp_j}} \times \frac{\measure(\pattern_j, \batch_i)}{\weight_\measure(\batch_i)} = \sum_{(\timestamp_i, \batch_i) \in \database} \frac{\measure(\pattern_j, \batch_i)\times \temporalbiasfunction{\dampingFactor}{\timestamp_n,\timestamp_i} }{\sum_{j\leq n}\weight_\measure(\batch_j)\times \temporalbiasfunction{\dampingFactor}{\timestamp_n,\timestamp_j}} = \frac{1}{\sum_{j\leq n}\weight_\measure(\batch_j)\times \temporalbiasfunction{\dampingFactor}{\timestamp_n,\timestamp_j}} \times (\sum_{(\timestamp_i, \batch_i) \in \database} \measure(\pattern_j, \batch_i)\times \temporalbiasfunction{\dampingFactor}{\timestamp_n,\timestamp_i} )$. From Definition \ref{def:patGlobUtilunderTemporalBias}, we have $\sum_{(\timestamp_i, \batch_i) \in \database} \measure(\pattern_j, \batch_i)\times \temporalbiasfunction{\dampingFactor}{\timestamp_n,\timestamp_i} =\globalUtilMeasure_\measure^{\dampingFactor}(\pattern_j, \database)$. 
 
    We can also deduce that $\sum_{j\leq n}\weight_\measure(\batch_j)\times \temporalbiasfunction{\dampingFactor}{\timestamp_n,\timestamp_j}=\sum_{\pattern \in \lang} \globalUtilMeasure_\measure^{\dampingFactor}(\pattern, \database)$ because both represent the sum of all pattern utilities under the utility measure \measure and the damping factor \dampingFactor. Hence, we get $\prob(\pattern_j=\reservoir[j], \database)= \frac{\globalUtilMeasure_\measure^{\dampingFactor}(\pattern_j, \database)}{\sum_{\pattern \in \lang} \globalUtilMeasure_\measure^{\dampingFactor}(\pattern, \database)}$ since $\sum_{j\leq n}\weight_\measure(\batch_j)\times \temporalbiasfunction{\dampingFactor}{\timestamp_n,\timestamp_j}=\sum_{\pattern \in \lang} \globalUtilMeasure_\measure^{\dampingFactor}(\pattern, \database)$.
\end{proof}
\end{sloppypar}

\paragraph{{Complexity analysis}}
Now, we assess the complexity of our approach by breaking it down into two points. Firstly, we show the complexity of inserting a pattern from an instance of the current batch. Secondly, we illustrate the complexity of drawing a pattern from a batch.

\noindent {\bf Complexity in number of insertions:} Because our approach is a sampling technique with replacement, a pattern occurrence from an instance of the current batch can be inserted at most \reservoirSize times into the reservoir \reservoir, with $\card{\reservoir}=\reservoirSize$. Without replacement, it has been shown that (Theorem 2 in \cite{efraimidis2006weighted}) after observing \datasize batches from a data stream where each of them has a strict positive weight, the number of reservoir insertions is $\complexity(\reservoirSize\times \log(\frac{\datasize}{\reservoirSize}))$. In our case, i.e, sampling with replacement, each  pattern occurrence can be inserted \reservoirSize  times into the reservoir. Therefore, if \myalgo is applied on a data stream, then the expected number of reservoir
insertions (without the initial \reservoirSize insertions) is in $\complexity(\reservoirSize^2\times \log(\frac{\card{\database}}{\reservoirSize}))$ regardless the pattern language (sequence or itemset). This theoretical result is reasonable because the reservoir size \reservoirSize is usually not high.

\noindent {\bf Complexity of the pattern selection step:} We focus on the pattern selection complexity which depends on the operator
$\sample{\lang}{*}{\measure}{\star}$. Let $\complexity(\sample{\lang}{*}{\measure}{\star})$ denote the complexity of this operator. For instance with Algorithm \ref{alg:op-sample}, it is used at each of \numberOfRealisations insertions, which leads to sample \numberOfRealisations patterns. In that case, the cost of \myalgo (Algorithm \ref{alg:op-sample}) in the pattern selection step is in $\complexity(\reservoirSize^2\times \log(\frac{\card{\database}}{\reservoirSize})) \times (\log(\card{\batch}_{\max}) + \complexity(\sample{\lang}{*}{\measure}{1}))$, with $\card{\batch}_{\max}$ the cardinality of the largest batch size of the visited ones. Then, the cost of \myalgo in the pattern selection step is $\complexity(\reservoirSize\times \log(\frac{\card{\database}}{\reservoirSize})) \times (\reservoirSize\times\log(\card{\batch}_{\max}) + \complexity(\sample{\lang}{*}{\measure}{\reservoirSize}))$.
In practice, the complexity of instance weighting $\complexity(\weight_\measure(\instance))$ and pattern drawing $\complexity(\sample{\lang}{*}{\measure}{1})$ relies on the pattern language $\lang$ and the interestingness measure $\measure$. Hence, \myalgo remains fast.

\section{Experimental Study}
\label{sec:Experiments}

\newcommand{\reutersEight}[0]{\texttt{Reuters8}\xspace}
\newcommand{\reutersFiftyTwo}[0]{\texttt{Reuters52}\xspace}
\newcommand{\Books}[0]{\texttt{Books}\xspace}
\newcommand{\webkb}[0]{\texttt{webkb}\xspace}
\newcommand{\cade}[0]{\texttt{cade}\xspace}
\newcommand{\retailTwentyTwentyThree}[0]{\texttt{Retail2023}\xspace}
\newcommand{\dOneMDixS}[0]{\texttt{D1M10S2T5I}\xspace}

\newcommand{\Susy}[0]{\texttt{Susy}\xspace}
\newcommand{\PowerC}[0]{\texttt{PowerC}\xspace}
\newcommand{\OnlineRetail}[0]{\texttt{ORetail}\xspace}
\newcommand{\KddcupNinetyNine}[0]{\texttt{kddcup99}\xspace}

\newcommand{\ECommerce}[0]{\texttt{ECommerce}\xspace}
\newcommand{\ChainstoreUI}[0]{\texttt{Chainstore}\xspace}
\newcommand{\FruithutUI}[0]{\texttt{Fruithut}\xspace}
\newcommand{\ChicagoCrimes}[0]{\texttt{ChicagoC}\xspace}

\newcommand{\labels}[0]{\ensuremath{{\cal C}}\xspace}

We evaluate the efficiency of \myalgo and the interest of the sampled patterns. More precisely, Section \ref{subsec:empirical_results} focuses on the speed of \myalgo with different batches and reservoir sizes as well as the maximal norm constraint. In Section \ref{subsec:accuracy_results}, in order to illustrate the usefulness of sampled patterns, we show how these patterns can be used to build associative classifiers dedicated to sequences from stream batches. 
All experiments are performed on a 2.71 GHz 2 Core CPU with 12 GB of RAM and the prototype of our method is implemented in Python. 
The minimum norm constraint is set to $\minNorm=1$ to avoid the empty set. 
Due to space limitations, we focus our experiments on sequential itemsets. In this case, we use the exponential decay with \alphaDecay=0.001 and maximal norm constraint $\maxNorm=10$.
However, Section \ref{subsec:speed_comp} compares \respat and \myalgo execution times for itemsets databases and more experiments with other pattern languages are available on the companion Github page.

\begin{table*}
    \centering
    {\scriptsize
    \caption{Average execution time per batch (in seconds) with different values of the damping factor ($\dampingFactor \in \{0.0, 0.1, 0.5\}$), the batch size ($\text{in } \{1000, 1500, 2000\}$) and the reservoir size ($\reservoirSize \in \{1000, 3000, 5000\}$)}
    \label{tab:time_per_batch}
        \begin{tabular}{|l|p{1.35cm}p{1.35cm}p{1.44cm}|p{1.44cm}p{1.44cm}p{1.35cm}|p{1.35cm}p{1.35cm}p{1.35cm}|}
        \cline{1-10}
         & \multicolumn{3}{c|}{Batch size 1000} & \multicolumn{3}{c|}{Batch size 1500} & \multicolumn{3}{c|}{Batch size 2000} \\ 
        \cline{2-10}
        Database & $\dampingFactor=0.0$ & $\dampingFactor=0.1$ & $\dampingFactor=0.5$ &
        $\dampingFactor=0.0$ & $\dampingFactor=0.1$ & $\dampingFactor=0.5$ &
        $\dampingFactor=0.0$ & $\dampingFactor=0.1$ & $\dampingFactor=0.5$ \\
        \hline
        \multicolumn{10}{c}{Reservoir size \reservoirSize=1000} \\
        \hline
        \Books & 0.36 $\pm$ 0.00 & 0.36 $\pm$ 0.00 & 0.36 $\pm$ 0.00 & 0.55 $\pm$ 0.01 & 0.56 $\pm$ 0.01 & 0.56 $\pm$ 0.01 & 0.70 $\pm$ 0.00 & 0.71 $\pm$ 0.00 & 0.72 $\pm$ 0.00 \\
        \reutersEight & 1.64 $\pm$ 0.01 & 1.65 $\pm$ 0.01 & 1.66 $\pm$ 0.03 & 2.57 $\pm$ 0.02 & 2.59 $\pm$ 0.02 & 2.60 $\pm$ 0.05 & 3.31 $\pm$ 0.01 & 3.34 $\pm$ 0.07 & 3.33 $\pm$ 0.06 \\
        \reutersFiftyTwo & 1.79 $\pm$ 0.05 & 1.78 $\pm$ 0.01 & 1.79 $\pm$ 0.01 & 2.79 $\pm$ 0.01 & 2.80 $\pm$ 0.02 & 2.79 $\pm$ 0.02 & 3.52 $\pm$ 0.05 & 3.58 $\pm$ 0.01 & 3.59 $\pm$ 0.02 \\
        \cade & 22.14 $\pm$ 0.12 & 22.26 $\pm$ 0.06 & 22.89 $\pm$ 0.24 & 34.84 $\pm$ 0.09 & 34.82 $\pm$ 0.18 & 35.73 $\pm$ 0.24 & 44.39 $\pm$ 0.14 & 44.42 $\pm$ 0.25 & 45.24 $\pm$ 0.41 \\
        \webkb & 18.14 $\pm$ 1.46 & 17.65 $\pm$ 1.87 & 19.18 $\pm$ 0.15 & 29.51 $\pm$ 0.27 & 28.52 $\pm$ 2.44 & 29.94 $\pm$ 0.37 & 38.28 $\pm$ 0.17 & 38.18 $\pm$ 0.24 & 38.26 $\pm$ 0.33 \\
        \hline

        \multicolumn{10}{c}{Reservoir size \reservoirSize=3000} \\
        \hline
        \Books & 0.35 $\pm$ 0.00 & 0.36 $\pm$ 0.00 & 0.37 $\pm$ 0.00 & 0.53 $\pm$ 0.00 & 0.54 $\pm$ 0.00 & 0.55 $\pm$ 0.00 & 0.70 $\pm$ 0.00 & 0.72 $\pm$ 0.00 & 0.78 $\pm$ 0.02 \\
        \reutersEight & 1.67 $\pm$ 0.01 & 1.68 $\pm$ 0.01 & 1.69 $\pm$ 0.04 & 2.52 $\pm$ 0.02 & 2.55 $\pm$ 0.06 & 2.52 $\pm$ 0.03 & 3.35 $\pm$ 0.03 & 3.65 $\pm$ 0.06 & 3.48 $\pm$ 0.10 \\
        \reutersFiftyTwo & 1.81 $\pm$ 0.03 & 1.80 $\pm$ 0.04 & 1.83 $\pm$ 0.02 & 2.73 $\pm$ 0.05 & 2.72 $\pm$ 0.04 & 2.70 $\pm$ 0.03 & 3.61 $\pm$ 0.02 & 3.62 $\pm$ 0.05 & 3.95 $\pm$ 0.22 \\
        \cade & 22.56 $\pm$ 0.15 & 22.84 $\pm$ 0.27 & 25.36 $\pm$ 0.99 & 34.05 $\pm$ 0.29 & 34.05 $\pm$ 0.34 & 36.36 $\pm$ 0.99 & 45.37 $\pm$ 0.54 & 48.46 $\pm$ 1.08 & 53.91 $\pm$ 5.66 \\
        \webkb & 26.51 $\pm$ 0.08 & 22.29 $\pm$ 6.22 & 27.43 $\pm$ 0.11 & 35.88 $\pm$ 7.20 & 36.37 $\pm$ 7.33 & 40.71 $\pm$ 0.12 & 54.44 $\pm$ 0.59 & 55.92 $\pm$ 0.59 & 56.99 $\pm$ 2.96 \\
        \hline

        \multicolumn{10}{c}{Reservoir size \reservoirSize=5000} \\
        \hline
        \Books & 0.35 $\pm$ 0.00 & 0.36 $\pm$ 0.00 & 0.38 $\pm$ 0.01 & 0.53 $\pm$ 0.00 & 0.54 $\pm$ 0.00 & 0.56 $\pm$ 0.00 & 0.71 $\pm$ 0.00 & 0.72 $\pm$ 0.00 & 0.74 $\pm$ 0.01 \\
        \reutersEight & 1.71 $\pm$ 0.01 & 1.71 $\pm$ 0.02 & 1.81 $\pm$ 0.02 & 2.56 $\pm$ 0.03 & 2.58 $\pm$ 0.03 & 2.60 $\pm$ 0.05 & 3.40 $\pm$ 0.03 & 3.39 $\pm$ 0.04 & 3.44 $\pm$ 0.11 \\
        \reutersFiftyTwo & 1.82 $\pm$ 0.01 & 1.84 $\pm$ 0.02 & 1.89 $\pm$ 0.04 & 2.73 $\pm$ 0.02 & 2.75 $\pm$ 0.07 & 2.80 $\pm$ 0.05 & 3.66 $\pm$ 0.05 & 3.70 $\pm$ 0.05 & 3.71 $\pm$ 0.07 \\
        \cade & 23.07 $\pm$ 0.26 & 23.59 $\pm$ 0.42 & 26.78 $\pm$ 0.82 & 34.52 $\pm$ 0.31 & 35.53 $\pm$ 0.96 & 38.88 $\pm$ 2.10 & 46.65 $\pm$ 0.85 & 46.78 $\pm$ 0.62 & 49.89 $\pm$ 1.14 \\
        \webkb & 35.13 $\pm$ 0.23 & 31.50 $\pm$ 8.98 & 29.36 $\pm$ 12.00 & 45.65 $\pm$ 12.32 & 28.87 $\pm$ 12.30 & 53.83 $\pm$ 0.39 & 71.80 $\pm$ 0.60 & 71.66 $\pm$ 0.58 & 71.67 $\pm$ 0.47 \\
\hline

        \end{tabular}
    }
\end{table*}

Table \ref{tab:databases} summarizes the characteristics of our benchmark of sequential databases. For each database, it shows the size $\card{\database}$, the number of distinct items $\card{\items}$, the maximal instance norm $\norm{\instance}_{\max}$, and the average norm of its instance $\norm{\instance}_{avg}$. The \Books sequential database is from SPMF\footnote{\url{https://www.philippe-fournier-viger.com/spmf/index.php}}, while the sequential databases \webkb, \reutersEight, \reutersFiftyTwo, and \cade are from Ana\footnote{\url{https://ana.cachopo.org/datasets-for-single-label-text-categorization}}.  Now we give some important comments for these datasets, which were adapted for the online classification task. For \webkb, \reutersEight, \reutersFiftyTwo, and \cade, we randomly merged the content of the train and test files of each dataset into a single file. For \Books, we also randomly merged the content of the files, and each instance is labeled by its author. In other words, we plan to predict the author of a given speech for the \Books dataset.

\begin{table}
\begin{center}
    {\small
	\caption{Statistics of benchmark sequential datasets}
	\label{tab:databases}
	\begin{tabular}{|l|rrrr|c|}
 
        \hline
        Database & \card{\database}& \card{\items} & $\norm{\instance}_{max}$ & $\norm{\instance}_{avg}$ & \card{\labels} \\
        \hline
        \hline
        \webkb & 4,168 & 7,770 & 20,628 & 133.36 & 4\\
        \reutersEight & 7,674& 22,931 & 533 & 64.53 & 8\\
        \reutersFiftyTwo & 9,100 & 25,611 & 595 & 68.62 & 52\\
        \cade & 40,983 &193,997 & 22,352& 116.44 & 12 \\
        \Books & 96,003 &14,452 & 379 & 24.04  & 10 \\
        \hline
	\end{tabular}
 }
\end{center}
\end{table}

\subsection{Evaluation of \myalgo speed on sequential data}
\label{subsec:empirical_results}

Table \ref{tab:time_per_batch} shows the behavior of \myalgo on different sequential databases as the reservoir size increases with different batch sizes. The experiments are repeated 5 times, and we can notice that the standard deviations are tiny. We see that the reservoir size has a slight impact on the execution time. Naturally, the batch size increases the execution time per batch. Let us notice that, for a given database with fixed size, the batch size has no impact on the global execution time. However, we note that the larger the database size, the higher the execution time. We also see that the damping factor (\dampingFactor) impacts the execution time; the higher it is, the higher the execution time due to numerous insertions. In all cases, we notice that \myalgo uses a reasonable execution time even with larger database such as \cade ($\approx 1,000 s$) while others take less than $200 s$.




\subsection{Accuracy of Sampling-Based Online Classification}
\label{subsec:accuracy_results}
The real problem that we face with these experiments is that, to the best of our knowledge, there is no state-of-the-art approach that can be used for online incremental learning in sequential data. All incremental classification approaches deal with itemsets. That is why we use cheater approaches from existing works on sequential data classification. In our framework, we use learning-duration, which is the length of the interval of timestamps in which our models learn based on the reservoir content. We also have the prediction-duration, which is the interval length for predicting only.

\begin{figure}
	\centering
	\includegraphics[scale=0.34]{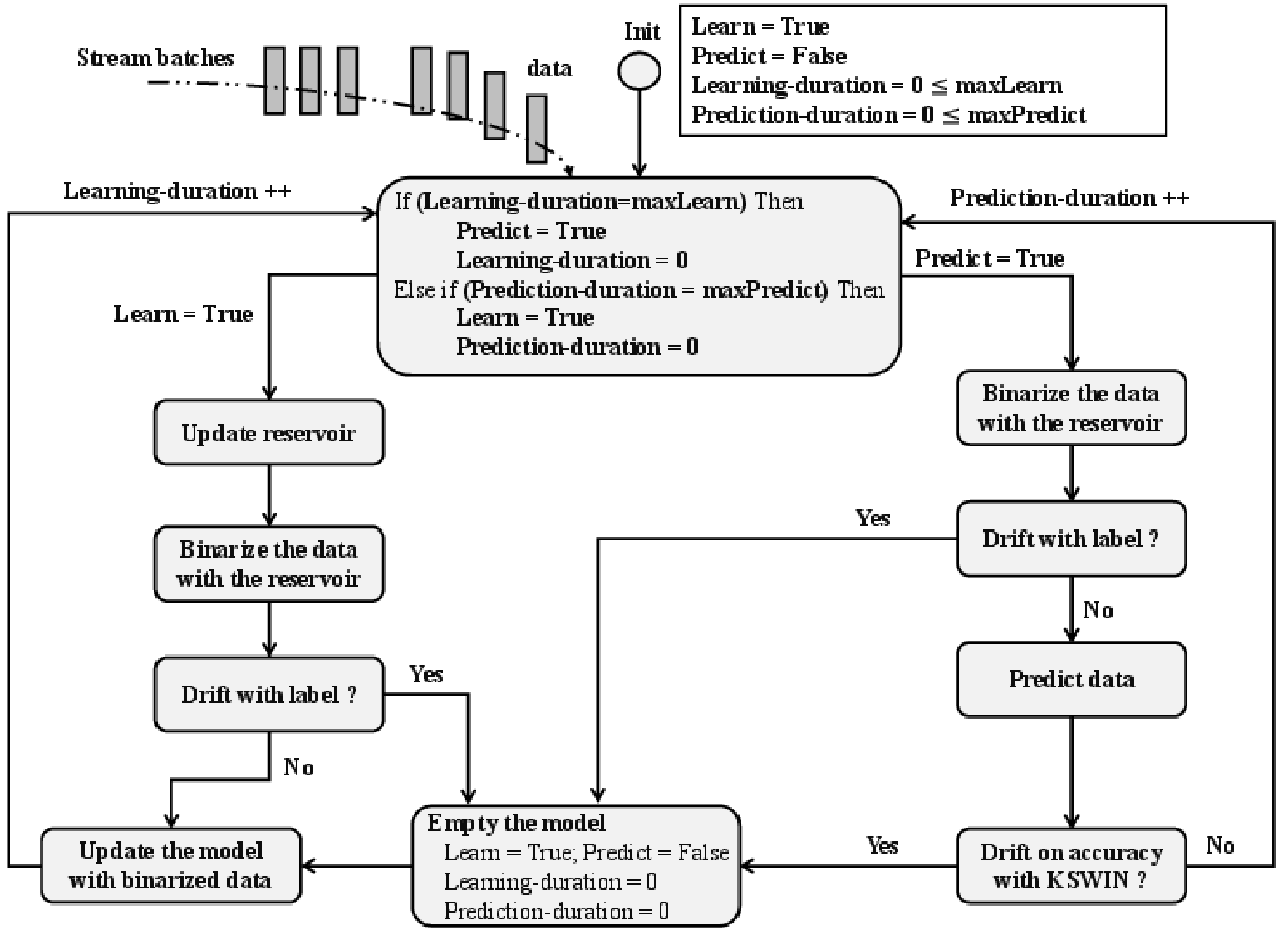}
 \caption{\myalgo-based classifier framework}
\label{fig:BSF_Classif_framework}
\end{figure}

As done in \cite{diopICDM18}, we represent each sequence \instance in a batch \batch as a tuple of $\reservoirSize+1$ values, where $d[j]=1$ if $\reservoir[j]\specialization \instance$ (0 otherwise) for $j \in [1..\reservoirSize]$, and $d[\reservoirSize+1]=c$, where $c$ is the class label of the sequence \instance and \reservoirSize is the reservoir size. Figure \ref{fig:BSF_Classif_framework} presents our learning classifier framework.

For the global accuracy, we use the average accuracy $\text{AvgAcc}$ for an incremental online classifier over multiple timestamps $\timestamp_0, \timestamp_1, \timestamp_2, \ldots, \timestamp_{n-1}$, with $\text{Acc}(\timestamp_i)$ is the accuracy at timestamp $\timestamp_i$, as: $
\text{AvgAcc} = \frac{1}{n} \sum_{i=0}^{n-1} \text{Acc}(\timestamp_i).$

\begin{table}
{\small
    \caption{Parameters for our adapted models}
    \label{tab:model_parameters}
    \centering
    \begin{tabular}{>{\raggedright\arraybackslash}p{0.7cm} >{\raggedright\arraybackslash}p{7.25cm}}
        \hline
        \textbf{Model} & \textbf{Parameters} \\
        \hline
        \MultinomialNB & alpha=0.0001 \\
        \hline
        \Perceptron & max\_iter=10000, tol=1e-3 \\
        \hline
        \PassiveAggressiveClassifier & max\_iter=1000, tol=1e-3, loss='hinge' \\
        \hline
        \MLPClassifier & hidden\_layer\_sizes=(500,100,), activation='identity', solver='adam', max\_iter=1000, warm\_start=False, random\_state=42 \\
        \hline
        \SGDClassifier & loss='hinge', penalty='l1', alpha=0.0001, max\_iter=1000, tol=1e-3 \\
        \hline
    \end{tabular}
}
\end{table}

In this section, we build many incremental learning classifiers for stream sequential data. In particular, based on the reservoir patterns under exponential decay to avoid the long tail issue, we show how to adapt machine learning models that have ``partial\_fit'' function such as \MLPClassifier(MLPClassifier \cite{Rumelhart1986LearningRB}), \SGDClassifier(SGDClassifier \cite{robbins1951stochastic}), \PassiveAggressiveClassifier(PassiveAggressiveClassifier \cite{Crammer06PassiveAggressiveClassifier}), \Perceptron(Perceptron \cite{Rosenblatt1958ThePerceptron}), and \MultinomialNB(MultinomialNB \cite{LewisMultinomialNB}) for incremental learning, where a model should be updated when drift (e.g., new labels, deteriorating accuracy) appears. For detecting deteriorating accuracy, we use the Kolmogorov-Smirnov test-based drift detector ``KSWIN (alpha=0.1)'' \cite{JMLRkswin} with default parameter settings.\footnote{We use the Scikit-learn package for the implementation \url{https://scikit-learn.org/stable/modules/linear_model.html}.}

\begin{figure}
	\centering
    {\small
	\includegraphics[scale=.355]{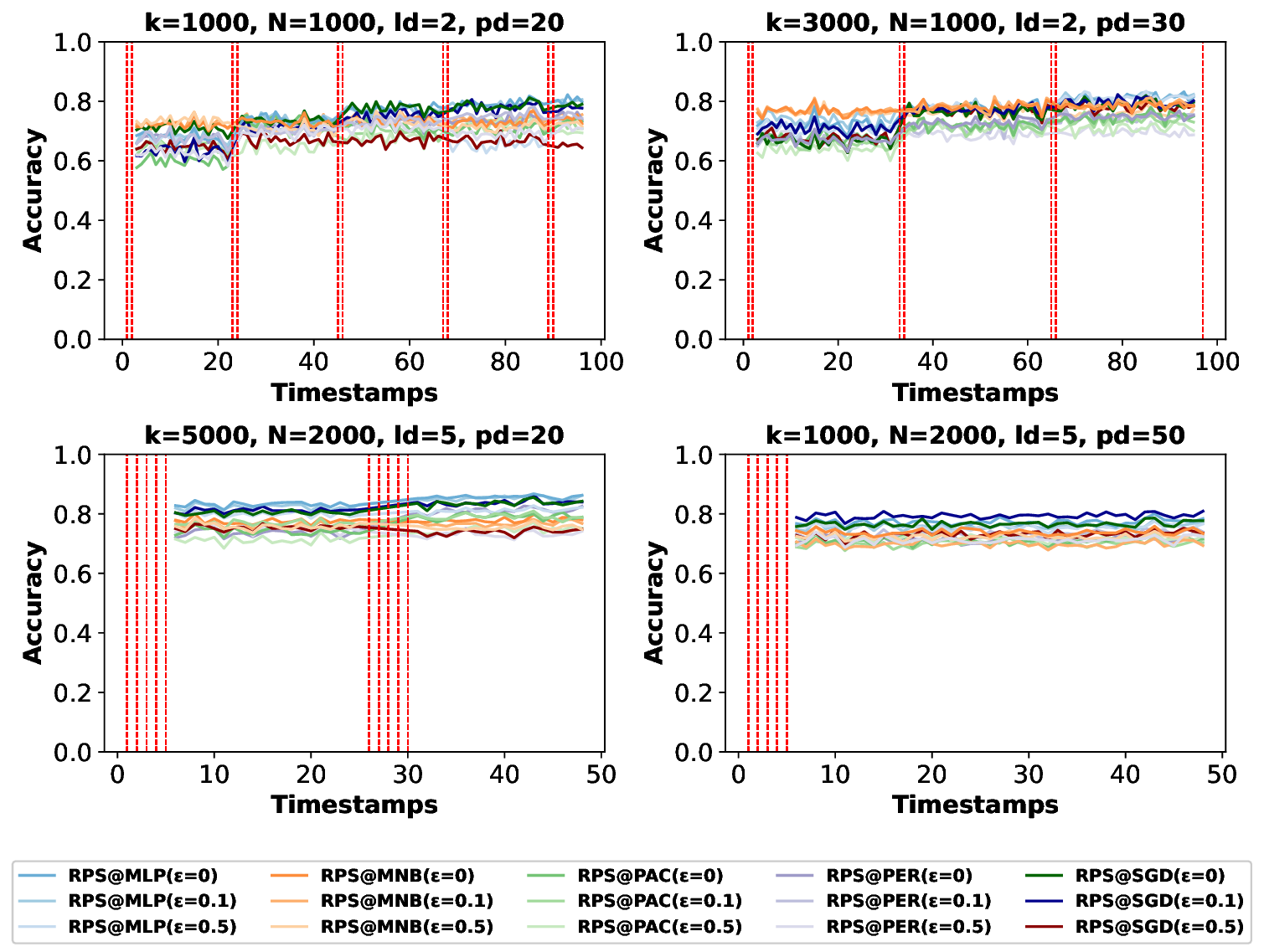}
    \caption{{\small Evolution of the accuracy per batch with different parameters on \Books. Learning timestamps are in red. \reservoirSize: reservoir size, $N$: batch size, $ld$: learning duration, $pd$: predict duration}}
	\label{fig:evolution_acc_per_batch_all_params_Books}
 }
\end{figure}

\paragraph{Impact of the \myalgo parameters on the accuracy}
We first present the behavior of \myalgo-based online classifiers with different parameter changes, such as the sample size, the reservoir size, the learning duration, and the predict duration, by considering the \Books database. Figure \ref{fig:evolution_acc_per_batch_all_params_Books} shows that many \myalgo-based models improve their performance after each learning interval. Additionally, the maintained reservoir patterns are representative enough to provide good accuracies for long-term predictions, thereby avoiding accuracy drift.

\begin{figure*}
	\centering
	\includegraphics[width=\textwidth, keepaspectratio]{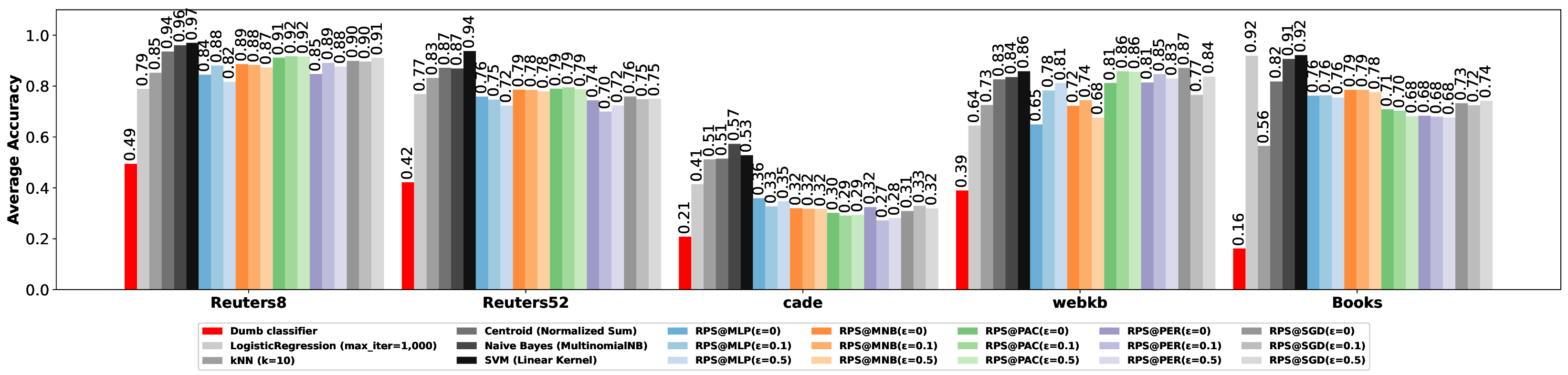}
 \caption{Comparison between \myalgo-based classifiers (with reservoir size \reservoirSize=10,000; batch size=1,000; learning duration=2 time-units, predict duration=52 time-units) vs cheater classifiers (with 50\% train and 50\% test)}
\label{fig:accuracy_comparison}
\end{figure*}

\paragraph{Accuracy comparison with cheater classifiers}
This section evaluates our approach by comparing it with cheater classifiers such as {\it Dumb classifier (strategy=`most\_frequent')}, {\it LogisticRegression (max\_iter=1,000)}, {\it KNN (k=10)}, {\it Centroid (Normalized Sum)}, {\it Naive Bayes (MultinomialNB)}, and {\it SVM (Linear Kernel)}. These classifiers are our references since they have access to all the training data (50\%), while \myalgo uses batches that appear during the learning intervals. This means that our goal is not to outperform them but to come closer to their performance as seen in Figure \ref{fig:accuracy_comparison}. However, \myalgo should be better than {\it Dumb classifier} which is a naturally lower baseline.

\myalgo demonstrates strong competitiveness with cheater methods. To enhance our approach for challenging databases like \cade, it is crucial to fine-tune both the model parameters and those of \myalgo. This includes adjusting aspects such as reducing the maximal norm, exploring alternative norm-based utility measures, and decreasing the rate of exponential decay.

\subsection{Execution time comparison between ResPat \cite{giacometti2021reservoir} and \myalgo on unweighted itemsets databases from the SPMF repository}
\label{subsec:speed_comp}

Table~\ref{tab:execution_times_comp} contains execution time comparisons between \respat \cite{giacometti2021reservoir}, and our approach, \myalgo. The experiments were repeated 5 times with different damping factors ($\dampingFactor \in \{0.0, 0.1, 0.5\}$), a batch size of $1,000$, a sample size of $k = 10,000$ without norm constraint (i.e., $M = \infty$). It shows that the \myalgo approach consistently outperforms \respat across all datasets and damping factors, with execution times up to $1,965$ times faster for ORetail and up to $1,774$ times faster for Kddcup99 at $\dampingFactor = 0$. Even for larger datasets like Susy, RPS maintains a significant speed advantage, being up to $398$ times faster at $\dampingFactor = 0$ and $25$ times faster at $\dampingFactor = 0.5$.

\begin{table*}[ht]
    \centering
    {\small
    \caption{Experimental results on the execution times without length constraint for {\bf \respat} and {\bf \myalgo}.}
    \label{tab:execution_times_comp}
    \begin{tabular}{|l|c|c|c|c|c|c|}
        \hline
        Database (size) & {\bf \respat}($\dampingFactor=0$) & {\bf \respat}($\dampingFactor=0.1$) & {\bf \respat}($\dampingFactor=0.5$) & {\bf \myalgo}($\dampingFactor=0$) & {\bf \myalgo}($\dampingFactor=0.1$) & {\bf \myalgo}($\dampingFactor=0.5$) \\
        \hline
        ORetail (541,909) & $1,474.08 \pm 11.49$ & $186.49 \pm 0.85$ & $343.42 \pm 0.12$ & $0.75 \pm 0.01$ & $4.24 \pm 0.03$ & $6.33 \pm 0.14$ \\
        Kddcup99 (1M) & $2,714.40 \pm 10.36$ & $1,478.57 \pm 13.03$ & $1,639.54 \pm 14.09$ & $1.53 \pm 0.02$ & $8.54 \pm 0.17$ & $14.05 \pm 0.31$ \\
        PowerC (1.04M) & $1,531.15 \pm 8.94$ & $399.04 \pm 7.97$ & $511.80 \pm 7.94$ & $1.35 \pm 0.01$ & $8.06 \pm 0.05$ & $12.91 \pm 0.11$ \\
        Susy (5M)  & $3,399.30 \pm 12.36$ & $1,879.02 \pm 17.03$ & $1,969.91 \pm 12.90$ & $8.53 \pm 0.18$ & $45.25 \pm 0.48$ & $77.86 \pm 6.08$ \\
        \hline
    \end{tabular}
    }
\end{table*}

\section{Conclusion}
\label{sec:Conclusion}
We introduced \myalgo, a novel reservoir pattern sampling approach for complex structured data in streams, such as sequential and weighted itemsets.
Our proposed method employs a multi-step technique that leverages the inverse incomplete Beta function and efficient computation of the normalization constant, resulting in a fast and effective pattern sampling approach.
Our extensive experiments demonstrate the robustness and versatility of \myalgo. Notably, we adapted several classification models for online sequential data classification with new labels, showing that sampled patterns significantly enhance the accuracy of online classifiers, achieving performance comparable to offline baselines. 

Future work will focus on extending \myalgo to graph streams, further broadening the applicability and impact of our research.
{\small
\bibliographystyle{IEEEtran}
\bibliography{RPS}
}
\end{document}